\documentclass[conference]{IEEEtran}
\usepackage{booktabs}
\IEEEoverridecommandlockouts

\usepackage{tabularray}
\usepackage{graphicx}

\usepackage{cite}
\usepackage{amsmath,amssymb,amsfonts}
\usepackage{algorithmic}
\usepackage{graphicx}
\usepackage{subcaption}
\usepackage{textcomp}
\usepackage[ruled]{algorithm2e}
\usepackage{booktabs}
\usepackage{bm}
\usepackage{array}
\usepackage{tabularray}
\usepackage{booktabs}
\usepackage{caption}
\usepackage{longtable}
\usepackage{multirow}

\usepackage{xcolor}
\def\BibTeX{{\rm B\kern-.05em{\sc i\kern-.025em b}\kern-.08em
    T\kern-.1667em\lower.7ex\hbox{E}\kern-.125emX}}
\begin{document}

\title{Lowering the Barrier of Machine Learning: Achieving Zero Manual Labeling in Review Classification Using LLMs\\


{\footnotesize \textsuperscript{*}Accepted to 2025 11th International Conference on Computing and Artificial Intelligence (ICCAI 2025)}}



\author{\IEEEauthorblockN{1\textsuperscript{st} Yejian Zhang}
\IEEEauthorblockA{\textit{Grad. School of Science and Technology} \\
\textit{Keio University}\\
Yokohama, Japan \\
zhangyejian@doi.ics.keio.ac.jp, ORCID: 0000-0002-6824-2589}
*Corresponding author
~\\
\and
\IEEEauthorblockN{2\textsuperscript{nd} Shingo Takada}
\IEEEauthorblockA{\textit{Grad. School of Science and Technology} \\
\textit{Keio University}\\
Yokohama, Japan \\
michigan@ics.keio.ac.jp}
~\\
}

\maketitle

\begin{abstract}
With the internet's evolution, consumers increasingly rely on online reviews for service or product choices, necessitating that businesses analyze extensive customer feedback to enhance their offerings. While machine learning-based sentiment classification shows promise in this realm, its technical complexity often bars small businesses and individuals from leveraging such advancements, which may end up making the competitive gap between small and large businesses even bigger in terms of improving customer satisfaction. This paper introduces an approach that integrates large language models (LLMs), specifically Generative Pre-trained Transformer (GPT) and Bidirectional Encoder Representations from Transformers (BERT)-based models, making it accessible to a wider audience. Our experiments across various datasets confirm that our approach retains high classification accuracy without the need for manual labeling, expert knowledge in tuning and data annotation, or substantial computational power. By significantly lowering the barriers to applying sentiment classification techniques, our methodology enhances competitiveness and paves the way for making machine learning technology accessible to a broader audience.
\end{abstract}


\section{Introduction}
\label{sec:introduction}

With the rapid development of the internet, online review platforms have become integral to consumer decision-making processes, providing a space for consumers to share experiences and opinions, thereby influencing others' purchasing decisions. Studies show that online reviews significantly affect consumer behavior. Chevalier and Mayzlin \cite{Chevalier2006} found that positive book reviews boost sales, while negative reviews deter buyers. Liu \cite{Liu2006} demonstrated that online reviews play a critical role in box office revenue. Hu, Pavlou, and Zhang \cite{Hu2008} highlighted the impact of review distribution on consumer perceptions. Dellarocas \cite{Dellarocas2003} discussed the promise and challenges of online review mechanisms, emphasizing how digitization has transformed consumer interaction. Forman, Ghose, and Wiesenfeld \cite{Forman2008} examined the influence of reviewer identity disclosure on sales. These studies underline the importance of online review platforms in shaping consumer behavior and business success.

Sentiment classification is a critical tool for businesses aiming to understand and respond to customer reviews effectively. By analyzing textual data from customer reviews, social media, and surveys, companies can gauge public sentiment toward their products, services, and overall brand. This insight enables businesses to make informed decisions, enhance customer satisfaction, and foster loyalty. For example, identifying common customer complaints allows businesses to address issues promptly, improving their service quality and customer retention rates \cite{sharma2023}. Additionally, sentiment analysis aids in brand management by monitoring social media platforms for real-time reviews, allowing companies to respond swiftly to negative comments and maintain a positive public image \cite{taboada2016, Bordoloi2023}.


Despite its benefits, the implementation of sentiment classification using machine learning presents several challenges:

1. \textbf{Resource-Intensive Training Data}: Creating an effective training dataset requires significant resources, including financial investment and domain-specific expertise. Accurate labeling of data is essential, and this often necessitates the involvement of professionals with specialized knowledge of the industry jargon and context \cite{9773849, Snow2008}.

2. \textbf{Technical Expertise in Machine Learning}: Applying machine learning techniques for sentiment classification demands high levels of technical expertise. Tasks such as fine-tuning algorithms, selecting appropriate models, and optimizing hyperparameters are complex and require deep understanding and experience. This expertise is often beyond the reach of small businesses and individuals, limiting their ability to leverage advanced sentiment analysis tools \cite{sharma2023}.

3. \textbf{Computational Resources}: The computational power needed to train and deploy machine learning models for sentiment classification is substantial. High-performance computing resources are necessary to handle large datasets and complex algorithms efficiently. This requirement further exacerbates accessibility issues for smaller entities, as they may not have the financial capability to invest in such infrastructure \cite{10433480, app13021003, Bordoloi2023}.

In conclusion, while sentiment classification offers significant advantages for businesses in understanding and responding to customer sentiments, the resource-intensive nature of its implementation poses a barrier to widespread adoption, particularly for smaller enterprises.

We have developed an approach comprised of LLM-based models, specifically the GPT-based Easy Sentiment Classification Startup GPT (ESCS-GPT) and the ALBERT/RoBERTa-based User Reviews Specific Language Model (URSLM), along with multiple classifiers to tackle the existing challenges in sentiment classification. In this framework, ESCS-GPT is utilized to generate a labeled training dataset, while URSLM is employed to obtain textual embeddings; additionally, we have integrated multiple classifiers to perform sentiment classification. This approach combines these modules to not only preserve high classification accuracy but also to eliminate the need for manual data labeling entirely. Importantly, our system does not require expert knowledge in tuning and data annotation or significant computational resources, thereby significantly lowering the barriers to employing sentiment classification techniques.

The main contributions of our approach are as follows:
\begin{enumerate}
    \item \textbf{High sentiment classification accuracy without any manual labeling:} Our approach combines ESCS-GPT and URSLMs along with multiple classifiers to achieve high sentiment classification accuracy without using any manually labeled data.
    \item \textbf{Reduction of required expertise:} Our approach does not require expert knowledge in machine learning or data annotation, making it accessible to a broader audience.
    \item \textbf{Low computational power requirements:} Our approach can efficiently operate under limited computational resources, further increasing its accessibility.
\end{enumerate}

The rest of this paper is organized as follows: 
Section \ref{Related Work} introduces related work. Section \ref{Proposed Approach} details the design of our approach. Section \ref{Implementation} discusses the implementation of our approach and the experimental setup. Section \ref{EVALUATION} presents the evaluation results. Section \ref{Threats to validity} addresses the threats to validity. Finally, Section \ref{Conclusion} concludes the paper and suggests directions for future work.


\section{Related Work}
\label{Related Work}

\subsection{Sentiment Classification Techniques}

Sentiment classification techniques have evolved significantly over the years, encompassing a range of methods from traditional machine learning to advanced deep learning approaches. Early techniques primarily relied on lexicon-based methods, which use predefined lists of sentiment-laden words to determine the sentiment polarity of a text. Taboada et al. \cite{taboada2011lexicon} describe these methods as simple and interpretable but often limited by contextual nuances and the need for extensive manual maintenance of lexicons.

Machine learning approaches have advanced sentiment classification. Pang et al. \cite{pang2002thumbs} and Sebastiani \cite{sebastiani2002machine} illustrate the use of Support Vector Machines (SVM), Naive Bayes, and logistic regression, which learn from labeled data to classify sentiment. These techniques benefit from handling a wide range of features extracted from text, such as n-grams and part-of-speech tags, but their performance heavily relies on the quality and size of the labeled training data.

In recent years, deep learning techniques have significantly improved sentiment classification. Kim \cite{kim2014convolutional} and Hochreiter and Schmidhuber \cite{hochreiter1997long} demonstrate the efficacy of Convolutional Neural Networks (CNNs) and Recurrent Neural Networks (RNNs), including Long Short-Term Memory (LSTM) networks. These models are effective at capturing complex syntactic and semantic patterns crucial for understanding sentiment in context. Transformer-based models like BERT \cite{devlin2018bert} have further advanced the field by enabling fine-tuning on specific sentiment analysis tasks with state-of-the-art performance.

Despite these advancements, challenges such as domain adaptation and computational resource requirements remain areas of active research. Combining multiple techniques, such as ensemble methods and hybrid approaches, as suggested by Zhang et al. \cite{zhang2018deep} and Yang et al. \cite{yang2019xlnet}, is a promising direction to enhance the robustness and accuracy of sentiment classification systems.

\subsection{Challenges in Manual Annotation and Machine Learning}
The process of manual annotation, essential for training supervised machine learning models, is not only labor-intensive and costly but also demands domain-specific expertise.

Deng et al. \cite{Deng2009} highlighted the substantial human effort required for image annotation in the development of the ImageNet database, which parallels the challenges faced in text annotation tasks. Similarly, Snow et al. \cite{Snow2008} evaluated annotations performed by non-experts for natural language tasks, revealing significant time investments and the necessity for annotators to possess both domain-specific knowledge and understanding of machine learning principles to ensure high-quality annotations.

\subsection{Background on Pre-trained Models}

The rapid advancement of pre-trained models has dramatically shifted the landscape of natural language processing, setting new benchmarks for what is achievable with modern AI technologies. Among the most notable advancements, Brown et al. \cite{Brown2020} introduced GPT-3, a state-of-the-art language model capable of performing various NLP tasks such as text completion, translation, and summarization, with minimal or no task-specific training data. This model is a prime example of zero-shot learning capabilities, where extensive pre-training on diverse datasets allows the model to generalize to new tasks.

While GPT can be applied in various tasks, its use scenarios often involve interactive text generation rather than direct application in structured tasks like large-scale classification, particularly in scenarios involving extensive volumes of user reviews. This usage tends to be computationally expensive and demands substantial resources, which can limit its practicality for continuous, large-scale deployment. Moreover, handling large volumes of text can lead to computational bottlenecks and efficiency issues due to the significant resource demands \cite{strubell-etal-2019-energy, chen2023frugalgptuselargelanguage, Narayanan2021, Rae2021, Brown2020}.

In contrast to the broad generative applications of GPT-3, BERT (Bidirectional Encoder Representations from Transformers), introduced by Devlin et al. \cite{devlin2018bert}, represents another milestone in the evolution of NLP. BERT advanced the understanding of the context in the text by processing words in relation to all other words in a sentence rather than one-directionally. This deeper understanding of context makes BERT especially effective when fine-tuned with an additional output layer, enabling it to perform well in a wide range of tasks, including text classification.

However, the deployment of BERT and its derivatives for direct classification tasks presents substantial challenges. These models typically require fine-tuning on labeled data to perform effectively in specific applications, addressing particular classification challenges \cite{devlin2018bert}. The fine-tuning process itself requires a deep understanding of machine learning workflows and the ability to optimize model parameters. Insights into these strategies are provided by Sun et al. \cite{sun2019finetune}, who discuss the nuances of fine-tuning BERT for text classification tasks. Moreover, Dodge et al. \cite{dodge2020finetune} explore the impact of various factors such as weight initializations and data orders on the fine-tuning of pre-trained language models like BERT. Additionally, Kovaleva et al. \cite{kovaleva2019secrets} examine the inner workings of BERT, underscoring the complexity of its behavior and the critical importance of thoroughly understanding its mechanisms to optimize performance effectively.

\section{Proposed Approach}

\label{Proposed Approach}
\subsection{Overview}
Previous approaches to sentiment classification often require extensive manual labeling, significant computational resources, and expert knowledge in machine learning. Our approach addresses these challenges by integrating LLM-based ESCS-GPT, LLM-based URSLM, and multiple classifiers.

To be specific, we propose the Easy Sentiment Classification Startup GPT (ESCS-GPT) to provide an intuitive and accessible method for users with limited expertise to generate training sets for downstream tasks. Since the primary vectorization and classification tasks are performed downstream, ESCS-GPT is essentially responsible for a lightweight initial classification task. This addresses the previously mentioned issues with GPT models, including high costs, significant computational demands, and stability concerns when handling intensive structured tasks. We also leveraged domain-specific knowledge in user reviews to further pre-train RoBERTa and ALBERT, resulting in the User Reviews Specific Language Models (URSLMs). These models generate embeddings tailored to user reviews, enhancing the accuracy of sentiment classification in downstream tasks. Finally, we integrated multiple machine learning classifiers to perform the sentiment classification tasks, further ensuring the robustness of our approach across different application scenarios.

\begin{figure}[]
  \centering
  \includegraphics[width=\linewidth]{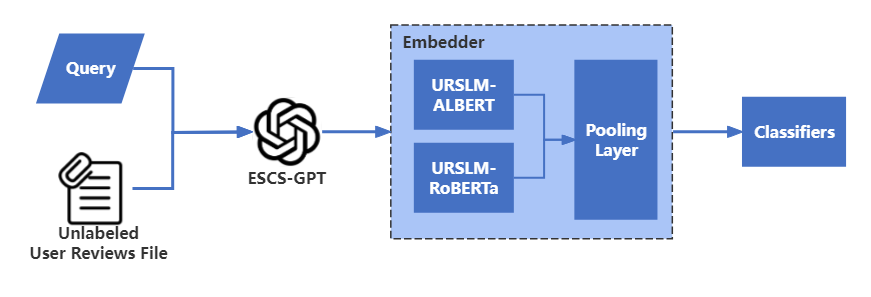}
  \caption{Overall workflow of our approach}
  \label{Fig1}
\end{figure}

In conclusion, our approach is designed to eliminate the need for manual labeling, expert knowledge in tuning and data annotation, and to reduce the reliance on extensive computational power. As a result, this enhanced accessibility allows a broader audience to benefit from advancements in machine learning, making high-performance sentiment classification more accessible. Figure \ref{Fig1} illustrates the overall workflow of our approach, and the subsequent subsections provide detailed explanations of each component.

\subsection{LLM-based Easy Sentiment Classification Startup GPT}
The Easy Sentiment Classification Startup GPT (ESCS-GPT) is an LLM-based conversational AI model configured using ChatGPT, built on the GPT-4 model. Specifically, through configuration commands (prompts), ESCS-GPT is instructed to select instances it deems most valuable for training downstream sentiment classifiers and to provide a labeled training set. The input to this module is user queries with unlabeled user reviews dataset as attachments, and the output is a labeled training set. This module is capable of generating a training set for downstream tasks through an intuitive chat format without requiring manual labeling and also reducing the need for coding expertise or domain knowledge.

\subsection{LLM-based User Review Specific Language Model}

The task of the User Review Specific Language Model (URSLM) is to convert user reviews into vectors for use as input to downstream classifiers. Specifically, URSLM leverages the capabilities of large language models (LLMs), including the bidirectional contextual understanding, general applicability, and transfer learning capabilities of BERT-based models, to generate high-quality text embeddings for downstream tasks.
We trained URSLMs based on RoBERTa-base and ALBERT-baseV2 using the domain corpus of user reviews for the Masked Language Model (MLM) task, which can be expressed by Formula (\ref{formulapretrain}). Additionally, we added a mean pooling layer to ensure that the embeddings represent the entire user review with a fixed length. By utilizing LLM-based URSLMs, we ensure that the generated embeddings are tailored to the domain-specific language of user reviews, thus enhancing the effectiveness of the downstream sentiment classification task.
 
\begin{equation}
\label{formulapretrain}
\mathcal{J}_{\text{MLM}} = \sum_{i=1}^{N} \mathbb{E}_{x_i \sim \mathcal{D}} \left[ \ell\left( \mathcal{M}(\text{mask}(x_i)), \text{original}(x_i) \right) \right]
\end{equation}
Where: \(\mathcal{J}_{\text{MLM}}\) represents the objective function during the Masked Language Modeling task. \(N\) is the total number of training samples. \(\mathbb{E}_{x_i \sim \mathcal{D}}\) denotes the expectation over the training data distribution \(\mathcal{D}\), where \(x_i\) is the input text from the dataset. \(\mathcal{M}\) represents the model (e.g., RoBERTa, ALBERT). \(\text{mask}(x_i)\) is the result of partially masking tokens in the input \(x_i\). \(\text{original}(x_i)\) are the original tokens at the masked positions, used for computing the loss function. \(\ell\) is the loss function, which measures the discrepancy between the model output for the masked input and the original tokens.


\subsection{Sentiment Classifiers} 
In the sentiment classification module, we implemented several classifiers, including Support Vector Machine (SVM), Decision Tree (DT), Random Forest (RF), and Logistic Regression (LR), to further ensure the robustness of our approach across different application scenarios. The input for this module consists of training and testing sets vectorized by the User Reviews Specific Language Models (URSLMs), and the output is the sentiment classification results (experimental results).

\section{Implementation}
\label{Implementation}
\subsection{Data preparation}
\begin{figure}[]
  \centering
  \includegraphics[width=0.7\linewidth]{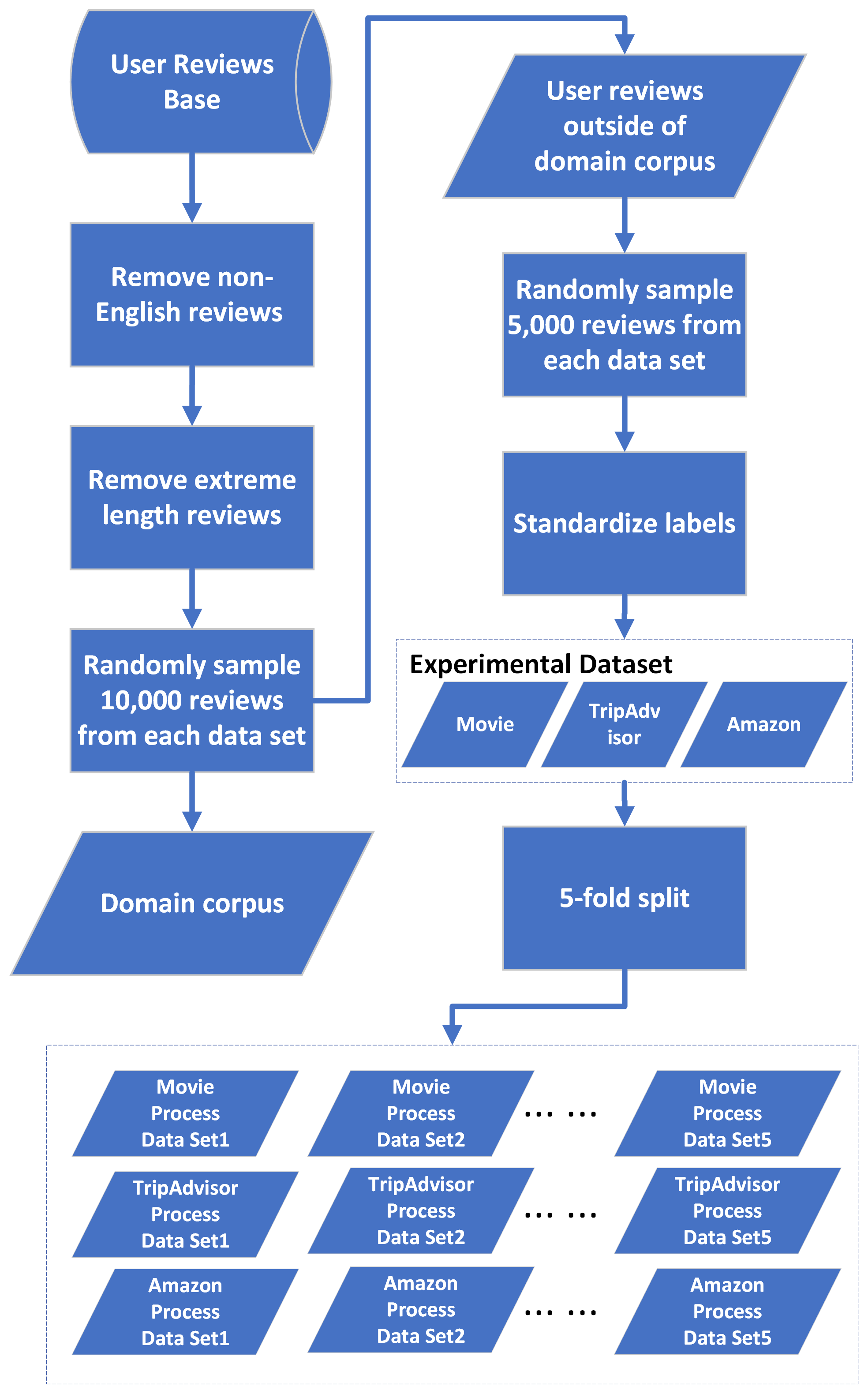}
  \caption{Workflow of data preparation}
  \label{Fig2}
\end{figure}

The data preparation process is divided into two parts: one is the Domain Corpus for further pre-training ALBERT and RoBERTa, and the other is the Experimental Dataset for evaluating our approach. The overall procedure of data preparation is shown in Figure \ref{Fig2}.

\begin{figure}[]
  \includegraphics[width=1\linewidth]{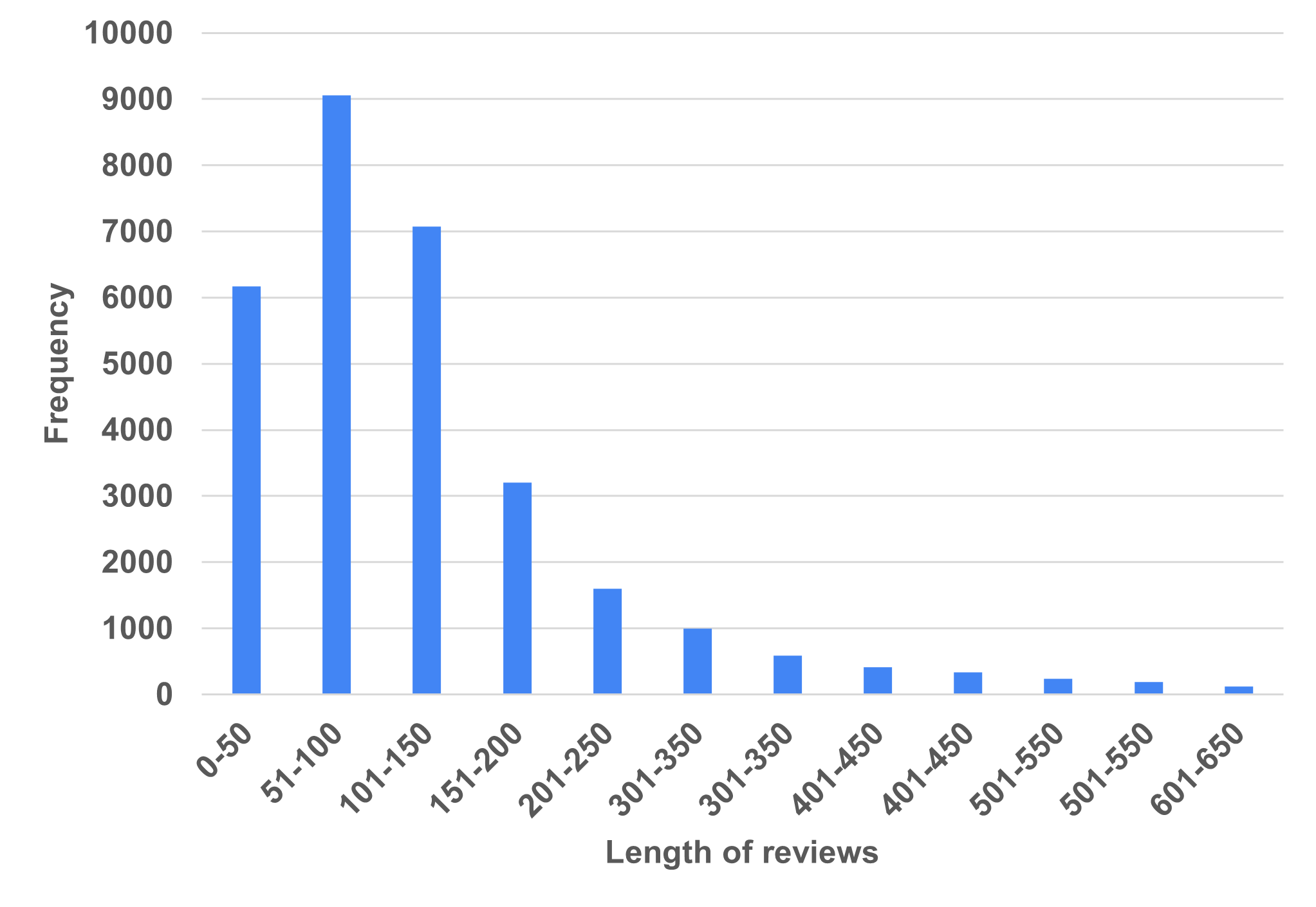}
  \caption{Length distribution of Domain Corpus}
  \label{Fig3}
\end{figure}

User Reviews Base contains the following three datasets: Amazon user reviews \cite{AmazonReviewsKaggle}, Movie user reviews \cite{maas2011learning}, and TripAdvisor user reviews\cite{bansal2018tripadvisor}.
The Amazon reviews dataset contains comprehensive user reviews in the retail industry, the TripAdvisor dataset represents user reviews in the service industry, and the Movie user reviews dataset contains user reviews in the cultural industry. We preprocessed each of the above three datasets by removing non-English reviews and removing reviews of extreme length. Specifically, we removed the top 5\% of the longest and shortest reviews in each dataset. After preprocessing, we randomly selected 10,000 user reviews in each dataset to form the Domain corpus for further pre-training ALBERT and RoBERTa. 30,000 user reviews were included in the Domain corpus, and the distribution of review lengths is shown in Figure \ref{Fig3}.

We then randomly sampled 5000 instances that were not chosen as the Domain Corpus from each of the three datasets. We standardized the labels of these sampled instances to form the Experimental Dataset. Specifically, for the Amazon and Movie datasets, the original annotations were binary labels indicating positive or negative sentiments. We used '0' to represent negative sentiment and '1' for positive sentiment. For the TripAdvisor dataset, which originally had ratings from 1 to 5, we adapted the data to fit the sentiment classification task by excluding reviews with a neutral rating (3) and labeling reviews with poor ratings (1-2) as negative sentiment ('0') and those with good ratings (4-5) as positive sentiment ('1').

\begin{figure}[]
  \centering
  \includegraphics[width=1\linewidth]{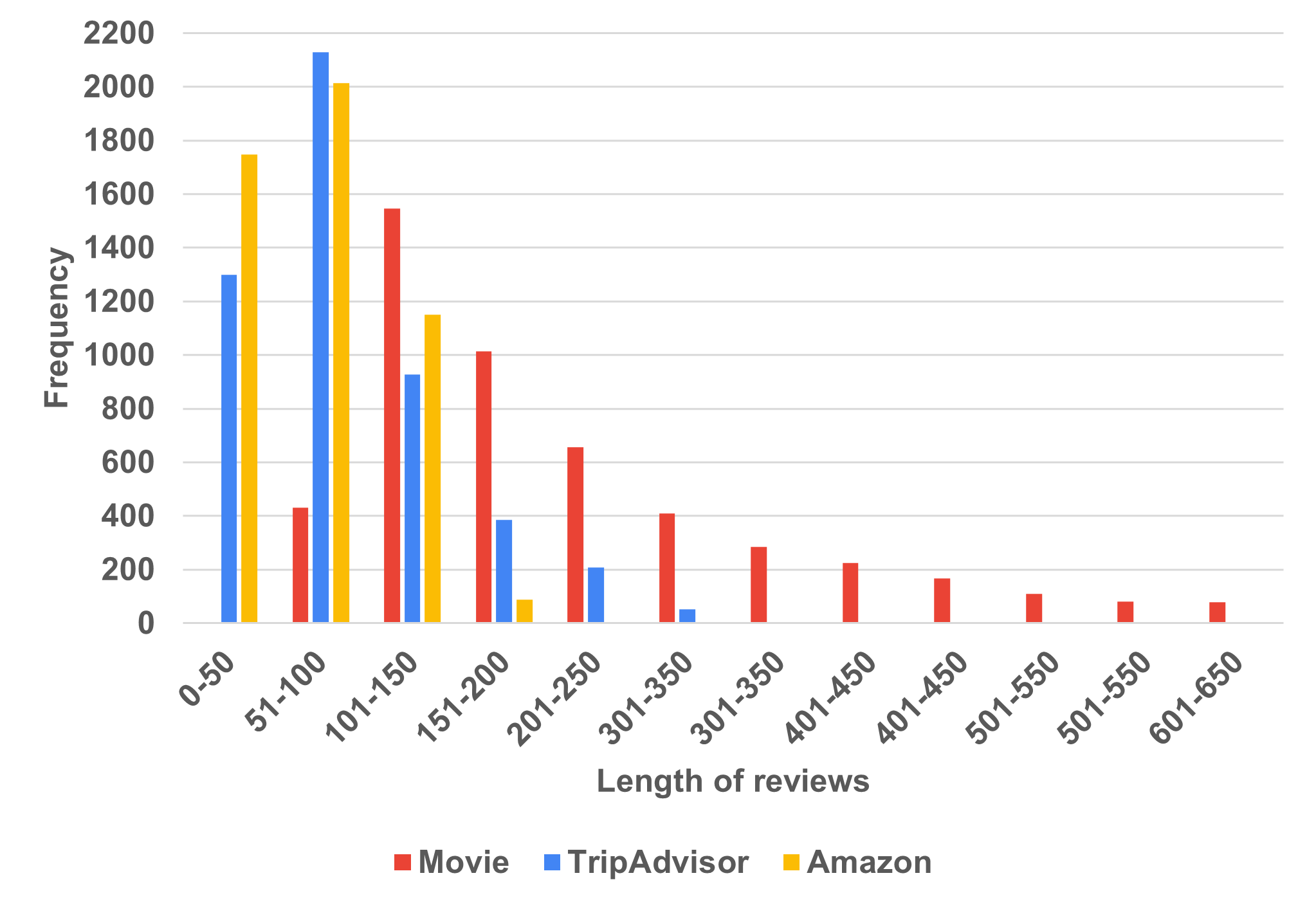}
  \caption{Length distribution across datasets}
  \label{Fig4}
\end{figure}

The length distributions of each dataset within the Experimental Dataset are illustrated in Figure \ref{Fig4}. Additionally, the distributions of positive and negative sentiments across these datasets are depicted in Table \ref{OLDFIG789}.


\begin{table}
\centering
\caption{Distribution of positive and negative sentiment labels in each dataset}
\label{OLDFIG789}
\scalebox{0.85}{
\resizebox{\columnwidth}{!}{%
\begin{tabular}{|l|l|l|} 
\hline
\textbf{Dataset}    & \textbf{Positive Labels (\%)} & \textbf{Negative Labels (\%)}  \\ 
\hline
Movie Reviews       & 50.4                          & 49.6                           \\ 
\hline
TripAdvisor Reviews & 83.0                          & 17.0                           \\ 
\hline
Amazon Reviews      & 49.3                          & 50.7                           \\
\hline
\end{tabular}%
}
}
\end{table}

\subsection{Experimental setup}
\subsubsection{Configuration of ESCS-GPT}

As described in Section III.B, we employed prompt engineering to develop the ChatGPT-based ESCS-GPT. The specific prompts are shown in Table \ref{tableprompt}. The design of our ESCS-GPT configuration commands (prompts), based on established principles in prompt engineering, including role assignment and detailed task descriptions, was refined through multiple iterations to ensure stable performance \cite{Liu2021, Gao2021, Schick2021}. Specifically, we first instructed the ESCS-GPT to act as an expert in machine learning and user reviews. Then, we required the ESCS-GPT to select instances it deems most valuable for training downstream sentiment classifiers and to provide a labeled training set. Additionally, since another primary goal of our ESCS-GPT is to allow users without relevant knowledge to directly obtain a usable training set through conversational interaction, we included instructions such as using a simple tool to ensure the ESCS-GPT can stably utilize ChatGPT's built-in capabilities to generate a training set without relying on external tools, thus ensuring accessibility. Finally, in the iterations, we added instructions such as balancing the dataset and considering comprehensively to ensure the stability of ESCS-GPT in the task of generating the training set.

\begin{table}
\centering
\caption{Configuration commands to setup ESCS-GPT}
\label{tableprompt}
\scalebox{0.8}{
\resizebox{\columnwidth}{!}{%
\begin{tblr}{
  hlines,
  hline{1,3} = {1}{0.08em},
}
\textbf{Configuration Commands}                                                                                                                                                                                                                                                                                                                                                                                                                                                                                                                                                                                                                                                                                                                                                                                                                                                                                                                                                                        \\
{You are an expert in machine learning and user reviews. \\Your task is to select the most valuable comments from \\customer-provided user reviews and provide sentiment \\labels to form a training set. Choose those comments \\that can most improve the performance of a sentiment \\classifier based on machine learning. The selected \\comments will be used for training downstream segment \\classifiers. When choosing comments, you should consider \\multiple factors, rather than simply selecting randomly or \\based solely on comment length. Also, avoid labeling the \\reviews only based on counting emotional words. In \\addition, please note the balance between each \\category of comments. Please use a simple tool \\for sentiment analysis to classify the reviews. \\Please comprehensively consider selecting instances\\that are most useful for training downstream classifiers, \\rather than random sampling.} 
\end{tblr}%
}
}
\end{table}

\subsubsection{Implementation of URSLMs \& Text Embedder}

\begin{figure}[]
  \centering
  \includegraphics[width=0.85\linewidth]{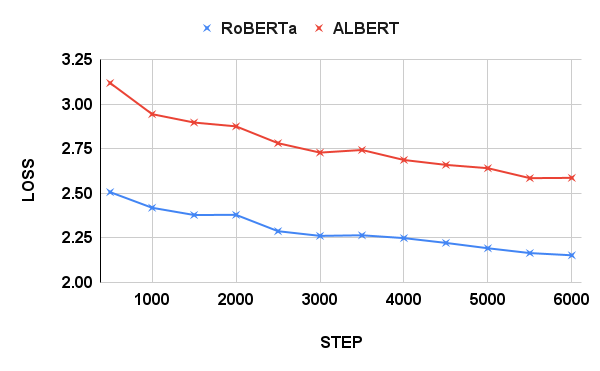}
  \caption{Training loss}
  \label{Fig10}
\end{figure}

Our approach's text embedder includes two types of User Reviews Specific Language Models (URSLMs) based on ALBERT-baseV2 and RoBERTa-base, along with a pooling layer. We conducted further pre-training on a domain-specific corpus. Specifically, we followed the Masked Language Model (MLM) procedure that involved masking random tokens in the training batches, computing the loss between predicted and actual tokens, and updating model parameters through forward and backward passes. This process was repeated for each batch across three training epochs, with a batch size of 14. Both RoBERTa-base and ALBERT-baseV2 were further pre-trained using this procedure, which allowed us to obtain URSLMs fine-tuned to our specific domain. During this further pre-training, the training loss for the MLM task was calculated using the cross-entropy loss function, shown in Formula (\ref{formulaloss}).

\begin{equation}
\label{formulaloss}
L = -\sum_{i \in M} \log(p_i(x_i))
\end{equation}
where \( M \) represents the set of masked positions in the input sequence, \( x_i \) is the actual token at the masked position \( i \), and \( p_i(x_i) \) is the predicted probability of the correct token \( x_i \) at the masked position. The training loss during further pre-training is shown in Figure \ref{Fig10}, which illustrates the convergence of the model over the training process. The training parameters for the language models are detailed in Table \ref{tableTrainLM}.

\begin{table}
\centering
\caption{Language modeling parameters}
\label{tableTrainLM}
\scalebox{0.75}{

\resizebox{\columnwidth}{!}{%
\begin{tabular}{|l|l|l|} 
\hline
                                   & \textbf{RoBERTa} & \textbf{ALBERT}  \\ 
\hline
\textbf{num\_train\_epochs}        & 3                & 3                \\ 
\hline
\textbf{batch\_size}               & 14               & 14               \\ 
\hline
\textbf{save\_steps}               & 500              & 500              \\ 
\hline
\textbf{max\_position\_embeddings} & 514              & 512              \\ 
\hline
\textbf{num\_attention\_heads}     & 12               & 12               \\ 
\hline
\textbf{num\_hidden\_layers}       & 12               & 12               \\ 
\hline
\textbf{hidden\_size}              & 768              & 768              \\
\hline
\end{tabular}%
}
}
\end{table}

Furthermore, we constructed the embedder by integrating a mean pooling layer using the Sentence Transformers framework, which allows the generation of fixed-length sentence embeddings that are effective for downstream tasks.

\subsubsection{Implementation of Sentiment Classifiers}
We implemented four classifiers: support vector machines (SVM), decision trees (DT), random forests (RF), and logistic regression (LR), which were chosen for their robustness and effectiveness in handling high-dimensional text data derived from user reviews. The SVM classifier employed the radial basis function (RBF) kernel due to its capability to handle complex patterns in text data. The decision tree model was configured with a maximum depth of 10 to help prevent overfitting, ensuring that the model remains generalizable across different datasets. To improve upon the stability and accuracy of decision trees, we also implemented a random forest classifier, an ensemble method that combines multiple decision trees to produce a more effective and robust model. Logistic regression (LR) was added for its ability to effectively handle high-dimensional text data through regularization methods. Each classifier was integrated into our sentiment classification pipeline, where they were tasked with classifying text embeddings generated by our further pre-trained URSLMs.

\section{EVALUATION}
\label{EVALUATION}
\subsection{Research Questions}
To evaluate our approach, we focus on the following research questions:
\begin{itemize}
\item RQ1: How does our approach perform across sentiment classification tasks on three distinct datasets from different domains, and which configuration of our analytical pipeline yields the best results?

\item RQ2: How much computational power does it take to use our approach?

\item RQ3: How does the performance of our approach compare to baseline methods that use manually labeled data?

\end{itemize}

\subsection{Evaluation Metrics}

The primary evaluation metric used was the accuracy score, and we also calculated the F1 score and recall. The formulas for calculating these metrics are given in Formula (\ref{ACC}), (\ref{Recall}), and (\ref{F1}). The confusion matrix used for these calculations is shown in Table \ref{Confusion}.

\begin{equation}
\label{ACC}
\text{Accuracy} = \frac{TP + TN}{TP + TN + FP + FN}
\end{equation}

\begin{equation}
\label{Recall}
\text{Recall} = \frac{TP}{TP + FN}
\end{equation}

\begin{equation}
\label{F1}
F1 = \frac{2 \times TP}{2 \times TP + FP + FN}
\end{equation}

\begin{table}[]
\centering
\caption{Confusion matrix}
\label{Confusion}
\scalebox{0.8}{

\resizebox{\columnwidth}{!}{%
\begin{tabular}{|l|l|l|} 
\hline
& \begin{tabular}[c]{@{}l@{}}\textbf{Predicted Positive}\\\textbf{ Sentiment}\end{tabular} & \begin{tabular}[c]{@{}l@{}}\textbf{Predicted Negative}\\\textbf{ Sentiment}\end{tabular}  \\ 
\hline
\begin{tabular}[c]{@{}l@{}}\textbf{Actual Positive}\\\textbf{ Sentiment}\end{tabular}  & TP                                                                                       & FN                                                                                        \\ 
\hline
\begin{tabular}[c]{@{}l@{}}\textbf{Actual Negative}\\\textbf{ Sentiment}\end{tabular} & FP                                                                                       & TN                                                                                        \\
\hline
\end{tabular}%
}
}
\end{table}

\subsection{Evaluation for RQ1}
We conducted 5-fold cross-validation on three datasets (Movie, TripAdvisor, Amazon). In each fold, the instances of the Process Data Set were provided as query attachments to ESCS-GPT after their labels were removed. ESCS-GPT was queried, as shown in Table \ref{GPT}, to generate a training set consisting of 100 labeled instances for training downstream classifiers. We then used URSLM-RoBERTa and URSLM-ALBERT to vectorize both the training set and the test set. These vectors were subsequently fed into the four classifiers, and the experimental results were calculated accordingly.

\begin{table}[h]
\caption{RQ1 ESCS-GPT query}
\label{GPT}
\centering
\begin{tabular}{p{0.8\linewidth}} 
\toprule
\textbf{ESCS-GPT query} \\
\midrule
Please carefully analyze the uploaded file of user reviews. Identify and select the 100 reviews that are most valuable for training our downstream sentiment classifier. For each selected review, assign a label of 'Positive' or 'Negative', use 1 for positive use 0 for negative. Please provide download links for the reviews you selected and annotated. \\
\bottomrule
\end{tabular}

\end{table}

The detailed 5-fold average performance of different pipelines on each dataset is summarized in Table \ref{RQ1TABNEW}. For the Movie dataset, ESCS+URSLM-RoBERTa +Logistic Regression achieved the best classification accuracy, and its classification accuracy was 83.4\% on average for 5-fold experiments. For the TripAdvisor dataset, ESCS+URSLM-ALBERT +RandomForest achieved the best classification accuracy, and its classification accuracy was 88.6\% on average for the 5-fold experiment. For the Amazon dataset, ESCS + URSLM-ALBERT +Logistic Regression achieved the best classification accuracy, and its classification accuracy was 80.8\% on average for the 5-fold experiment.
This demonstrates that our proposed approach that combines ESCS-GPT, URSLM, and multiple classifiers achieves high classification accuracy across various sentiment classification scenarios, reflecting the effectiveness and robustness of our approach.

\begin{table*}
\centering
\caption{RQ1 performance of our approach on each dataset}
\label{RQ1TABNEW}
\scalebox{1}{
\resizebox{\textwidth}{!}{%
\begin{tblr}{
  cell{1}{1} = {r=2}{},
  cell{1}{2} = {c=3}{c},
  cell{1}{5} = {c=3}{c},
  cell{1}{8} = {c=3}{c},
  vline{2-3,6} = {1}{},
  vline{5,8} = {2}{},
  vline{2,5,8} = {1-10}{},
  hline{1,11} = {-}{0.08em},
  hline{2} = {2-10}{},
  hline{3} = {-}{},
}
\textbf{Model}              & \textbf{Movie}    &                   &                 & \textbf{TripAdvisor} &                   &                 & \textbf{Amazon}   &                   &                 \\
                            & \textbf{Accuracy} & \textbf{F1 Score} & \textbf{Recall} & \textbf{Accuracy}    & \textbf{F1 Score} & \textbf{Recall} & \textbf{Accuracy} & \textbf{F1 Score} & \textbf{Recall} \\
\textbf{ESCS+URSLM-RoBERTa +SVM} & 0.767±0.02 & 0.746±0.03 & 0.684±0.06 & 0.650±0.01  & 0.738±0.01 & 0.595±0.02 & 0.650±0.02 & 0.569±0.04 & 0.469±0.04 \\
\textbf{ESCS+URSLM-RoBERTa +DT}  & 0.702±0.02 & 0.686±0.03 & 0.647±0.06 & 0.823±0.02  & 0.887±0.02 & 0.838±0.02 & 0.670±0.02 & 0.661±0.06 & 0.668±0.11 \\
\textbf{ESCS+URSLM-RoBERTa +RF}  & 0.829±0.01 & 0.826±0.01 & 0.803±0.03 & 0.869±0.02  & 0.915±0.01 & 0.853±0.02 & 0.789±0.02 & 0.753±0.04 & 0.657±0.05 \\
\textbf{ESCS+URSLM-RoBERTa +LR}  & 0.834±0.01 & 0.833±0.01 & 0.819±0.02 & 0.701±0.02  & 0.782±0.01 & 0.646±0.02 & 0.746±0.03 & 0.672±0.06 & 0.538±0.07 \\
\textbf{ESCS+URSLM-ALBERT +SVM}  & 0.811±0.01 & 0.791±0.01 & 0.707±0.02 & 0.873±0.02  & 0.918±0.01 & 0.856±0.02 & 0.802±0.01 & 0.758±0.03 & 0.634±0.04 \\
\textbf{ESCS+URSLM-ALBERT +DT}   & 0.689±0.03 & 0.687±0.03 & 0.682±0.06 & 0.785±0.06  & 0.857±0.05 & 0.791±0.09 & 0.651±0.02 & 0.644±0.04 & 0.648±0.08 \\
\textbf{ESCS+URSLM-ALBERT +RF}   & 0.813±0.01 & 0.814±0.01 & 0.813±0.03 & 0.886±0.02  & 0.927±0.01 & 0.877±0.02 & 0.795±0.02 & 0.764±0.03 & 0.679±0.05 \\
\textbf{ESCS+URSLM-ALBERT +LR}   & 0.825±0.01 & 0.826±0.01 & 0.827±0.02 & 0.874±0.02  & 0.919±0.01 & 0.858±0.02 & 0.808±0.01 & 0.778±0.03 & 0.687±0.05 

\end{tblr}%
}
}
\end{table*}

\subsection{Evaluation for RQ2}

To answer RQ2, we set up the environment using freely available resources on Colab \cite{GoogleColab}, which is accessible to any registered user. The system configuration consisted of two Intel(R) Xeon(R) CPUs operating at 2.20GHz, 12GB of CPU RAM, and a Tesla T4 GPU with 15GB of GPU RAM. We ran the 5-fold experiments described in RQ1 within this environment. During these experiments, the maximum system CPU memory usage was 2.26GB and the maximum system GPU memory usage was 2.92GB, and the running time of each pipeline is shown in Table \ref{RunTime}. The results indicate that our approach can be efficiently executed on widely available free cloud platforms, which are often limited in computational resources. The observed resource usage, including modest CPU and GPU memory consumption and short pipeline execution times, reflects the computational efficiency of our method, highlighting its accessibility to a broader audience. This efficient utilization of resources further underscores the practicality and suitability of the approach in scenarios with constrained computational environments.

\begin{table}[]
\caption{RQ2 running time (seconds) of each pipeline}
\label{RunTime}
\centering
\scalebox{0.9}{
\resizebox{\columnwidth}{!}{%
\begin{tblr}{
  cell{2}{1} = {r=3}{},
  cell{5}{1} = {r=3}{},
  cell{8}{1} = {r=3}{},
  cell{11}{1} = {r=3}{},
  cell{14}{1} = {r=3}{},
  cell{17}{1} = {r=3}{},
  cell{20}{1} = {r=3}{},
  cell{23}{1} = {r=3}{},
  hline{1-2} = {-}{},
  hline{5,8,11,14,17,20,23,26} = {1-5}{},
}
\textbf{Model}                                          & \textbf{Data Set} & {\textbf{Average }\\\textbf{Vectorization}\\\textbf{Time}} & {\textbf{Average}\\\textbf{Training}\\\textbf{Time}} & {\textbf{Average}\\\textbf{Prediction}\\\textbf{Time}} \\
{\textbf{ESCS+URSLM}\\\textbf{-RoBERTa}\\\textbf{+SVM}} & Movie             & 20.9428                                                    & 0.0045                                               & 0.1239                                                 \\
                                                        & TripAdvisor       & ~ 9.5239                                                   & 0.0042                                               & 0.0586                                                 \\
                                                        & Amazon            & ~ 7.3724                                                   & 0.0036                                               & 0.0474                                                 \\
{\textbf{ESCS+URSLM}\\\textbf{-RoBERTa}\\\textbf{+DT}}  & Movie             & 20.9428                                                    & 0.0402                                               & 0.0008                                                 \\
                                                        & TripAdvisor       & ~ 9.5239                                                   & 0.0157                                               & 0.0007                                                 \\
                                                        & Amazon            & ~ 7.3724                                                   & 0.0193                                               & 0.0006                                                 \\
{\textbf{ESCS+URSLM}\\\textbf{-RoBERTa}\\\textbf{+RF}}  & Movie             & 20.9428                                                    & 0.3566                                               & 0.0195                                                 \\
                                                        & TripAdvisor       & ~ 9.5239                                                   & 0.1855                                               & 0.0138                                                 \\
                                                        & Amazon            & ~ 7.3724                                                   & 0.1709                                               & 0.0142                                                 \\
{\textbf{ESCS+URSLM}\\\textbf{-RoBERTa }\\\textbf{+LR}} & Movie             & 20.9428                                                    & 0.0545                                               & 0.0028                                                 \\
                                                        & TripAdvisor       & ~ 9.5239                                                   & 0.0259                                               & 0.0022                                                 \\
                                                        & Amazon            & ~ 7.3724                                                   & 0.0471                                               & 0.0018                                                 \\
{\textbf{ESCS+URSLM}\\\textbf{-ALBERT }\\\textbf{+SVM}} & Movie             & 25.2097                                                    & 0.0035                                               & 0.0478                                                 \\
                                                        & TripAdvisor       & 11.4365                                                    & 0.0055                                               & 0.0597                                                 \\
                                                        & Amazon            & ~ 8.9138                                                   & 0.0037                                               & 0.0879                                                 \\
{\textbf{ESCS+URSLM}\\\textbf{-ALBERT}\\\textbf{+DT}}   & Movie             & 25.2097                                                    & 0.0178                                               & 0.0006                                                 \\
                                                        & TripAdvisor       & 11.4365                                                    & 0.0231                                               & 0.0008                                                 \\
                                                        & Amazon            & ~ 8.9138                                                   & 0.0336                                               & 0.0007                                                 \\
{\textbf{ESCS+URSLM}\\\textbf{-ALBERT}\\\textbf{+RF}}   & Movie             & 25.2097                                                    & 0.1989                                               & 0.0134                                                 \\
                                                        & TripAdvisor       & 11.4365                                                    & 0.3102                                               & 0.0188                                                 \\
                                                        & Amazon            & ~ 8.9138                                                   & 0.3043                                               & 0.0247                                                 \\
{\textbf{ESCS+URSLM}\\\textbf{-ALBERT}\\\textbf{+LR}}   & Movie             & 25.2097                                                    & 0.0424                                               & 0.0026                                                 \\
                                                        & TripAdvisor       & 11.4365                                                    & 0.0635                                               & 0.0042                                                 \\
                                                        & Amazon            & ~ 8.9138                                                   & 0.0793                                               & 0.0034                                                 
\end{tblr}%
}
}
\end{table}

\subsection{Evaluation for RQ3}

To conduct comparative experiments, we set up two types of baseline models: classic baseline models and baseline models using LLMs. The classic baseline models include TF-IDF combined with classifiers, specifically Support Vector Machine (SVM), Decision Tree (DT), Random Forest (RF), and Logistic Regression (LR). Additionally, we used Bag of Words (BoW) features with the same set of classifiers. For the LLM-based baseline models, we utilized RoBERTa-base and ALBERT-baseV2, combined with the aforementioned classifiers. We employed the same Experimental Dataset for 5-fold cross-validation. Due to the absence of the ESCS-GPT component, all of these models required manually annotated training sets. Therefore, for each fold, we randomly sampled 100 instances from the Process Data Set and used the original dataset's labels. To further ensure fairness in the comparative experiments, we performed ten random samplings for each fold and calculated the average results.

\begin{figure*}[]
  \centering
  \includegraphics[width=1\linewidth]{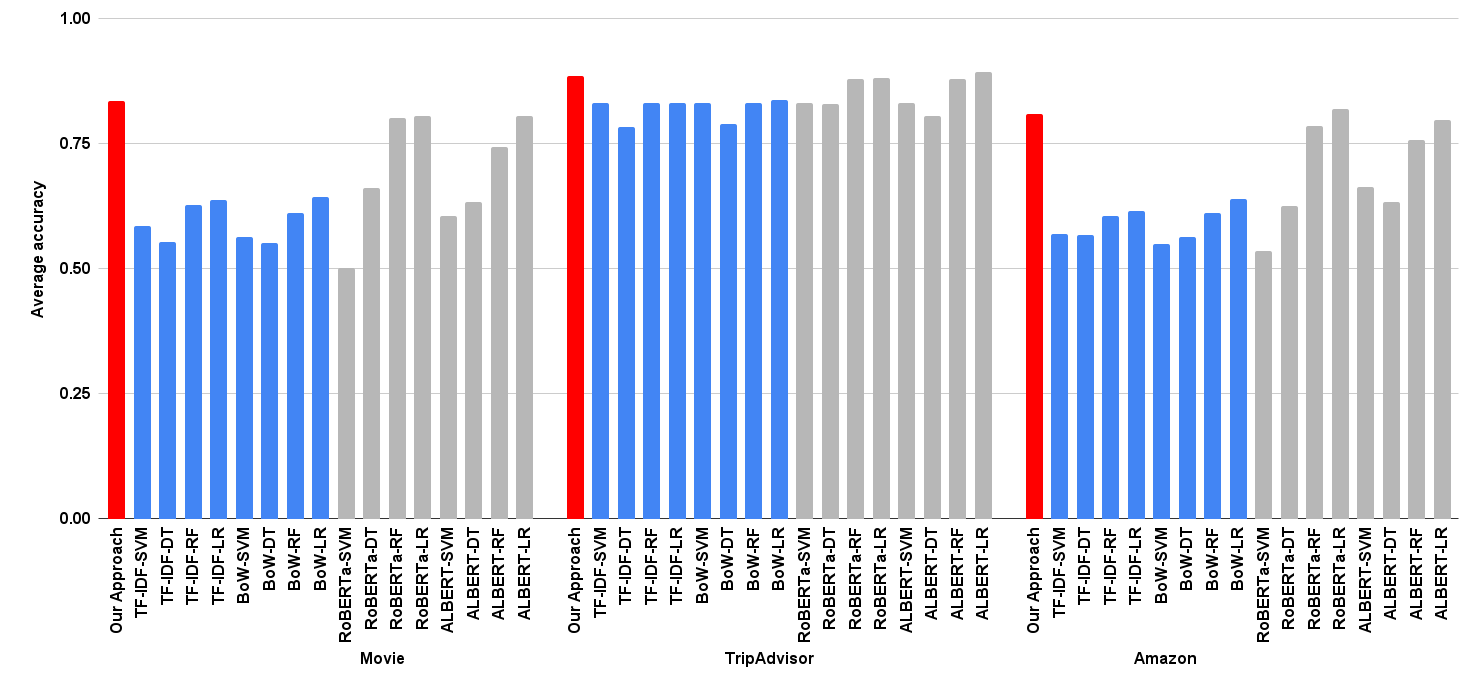}
  \caption{RQ3 comparative experiment results}
  \label{RQ3Fig}
\end{figure*}

The experimental results are presented in Figure \ref{RQ3Fig}, where the red bars in each dataset represent the average accuracy of the best-performing pipeline in our approach, the blue bars represent the average accuracies of the classic baseline models, and the gray bars represent the average accuracies of the LLM-based baseline models. Detailed performance of baseline models is provided in Table \ref{RQ3Results}. On the Movie dataset, the best classification accuracy for the classic baseline models was 64.2\%, whereas the best accuracy for the LLMs baseline models was 80.6\%. On the TripAdvisor dataset, the best accuracy for the classic baseline models was 83.7\%, for the LLMs baseline models was 89.4\%. On the Amazon dataset, the classic baseline models' best accuracy was 63.8\%, and the best accuracy of the LLMs baseline models was 81.8\%.


\begin{table*}[]
\centering
\caption{RQ3 detailed results of baseline models}
\label{RQ3Results}

\renewcommand{\arraystretch}{0.1}
\scalebox{1}{

\resizebox{\textwidth}{!}{%

\begin{tblr}{
  cell{1}{1} = {c=4}{c},
  cell{1}{5} = {c=4}{c},
  cell{1}{9} = {c=4}{c},
  vline{2,6} = {1}{},
  vline{5,9} = {1-10}{},
  hline{1,11} = {-}{0.08em},
  hline{2,3} = {-}{},
}
\textbf{Movie}  &                       &                 &                       & \textbf{TripAdvisor} &                       &                 &                       & \textbf{Amazon} &                       &                 &                       \\
\textbf{Method} & \textbf{Acc±STD} & \textbf{Method} & \textbf{Acc±STD} & \textbf{Method}      & \textbf{Acc±STD} & \textbf{Method} & \textbf{Acc±STD} & \textbf{Method} & \textbf{Acc±STD} & \textbf{Method} & \textbf{Acc±STD} \\
TFIDF-SVM & 0.584±0.08   & RoBERTa-SVM & 0.501±0.00   & TFIDF-SVM & 0.830±0.00   & RoBERTa-SVM & 0.830±0.00   & TFIDF-SVM & 0.568±0.06   & RoBERTa-SVM & 0.534±0.07    \\
TFIDF-DT  & 0.553±0.04   & RoBERTa-DT  & 0.661±0.04   & TFIDF-DT  & 0.783±0.02   & RoBERTa-DT  & 0.830±0.02   & TFIDF-DT  & 0.568±0.02   & RoBERTa-DT  & 0.624±0.03    \\
TFIDF-RF  & 0.626±0.05   & RoBERTa-RF  & 0.802±0.03   & TFIDF-RF  & 0.830±0.00   & RoBERTa-RF  & 0.878±0.01   & TFIDF-RF  & 0.605±0.02   & RoBERTa-RF  & 0.786±0.02    \\
TFIDF-LR  & 0.638±0.08   & RoBERTa-LR  & 0.806±0.03   & TFIDF-LR  & 0.830±0.00   & RoBERTa-LR  & 0.881±0.01   & TFIDF-LR  & 0.615±0.05   & RoBERTa-LR  & 0.818±0.01    \\
BoW-SVM   & 0.564±0.04   & ALBERT-SVM  & 0.606±0.09   & BoW-SVM   & 0.830±0.00   & ALBERT-SVM  & 0.830±0.00   & BoW-SVM   & 0.550±0.04   & ALBERT-SVM  & 0.663±0.05    \\
BoW-DT    & 0.551±0.03   & ALBERT-DT   & 0.632±0.03   & BoW-DT    & 0.790±0.02   & ALBERT-DT   & 0.805±0.03   & BoW-DT    & 0.562±0.03   & ALBERT-DT   & 0.632±0.02    \\
BoW-RF    & 0.611±0.06   & ALBERT-RF   & 0.744±0.03   & BoW-RF    & 0.830±0.00   & ALBERT-RF   & 0.878±0.00   & BoW-RF    & 0.610±0.04   & ALBERT-RF   & 0.757±0.02    \\
BoW-LR    & 0.642±0.03   & ALBERT-LR   & 0.805±0.01   & BoW-LR    & 0.837±0.01   & ALBERT-LR   & 0.894±0.01   & BoW-LR    & 0.638±0.02   & ALBERT-LR   & 0.797±0.02

\end{tblr}%
}}
\end{table*}

In conclusion, even without using any manually labeled data, our approach significantly outperformed the classic baseline models on the Movie and Amazon datasets and performed better than the classic baseline models on the TripAdvisor dataset. For the comparative models utilizing LLMs, our approach's performance either surpassed or closely approximated the results of the comparative experiments that used professionally labeled data from the original datasets.

\section{Threats to validity}
\label{Threats to validity}

The threats to validity in our study can be classified into construct, internal, and external validity concerns. For construct validity, the accuracy of annotations in the experimental data is a key threat. To mitigate this, we selected reliable datasets: Amazon and TripAdvisor reviews from Kaggle, both with perfect usability scores, and the movie reviews dataset from Stanford University, specifically designed for sentiment classification. We also addressed configuration command issues by referring to relevant prompt engineering studies and conducting iterative testing to ensure stability. For internal validity, the primary threat is overfitting. We mitigated this by ensuring no overlap between the training data for BERT-based models and the data used for sentiment classification experiments. K-fold cross-validation and average metrics across experiments were used to assess consistency and reduce overfitting. The main threat to external validity is the generalizability of our model. We addressed this by selecting datasets from three distinct domains—retail, service, and cultural industries—reflecting diverse user interactions. While the use of one dataset per domain limits generalizability, leveraging these diverse datasets enhances robustness. Future work will incorporate additional datasets to further validate the results. Additionally, using LLMs carries a risk of inheriting biases from pre-training data, which we mitigated through domain-specific fine-tuning to better align the URSLM with the target domain.

\section{Conclusion}
\label{Conclusion}
In this study, we introduced a methodology for sentiment classification that significantly reduces the barriers to utilizing machine learning for sentiment classification. We proposed ESCS-GPT and URSLM based on Generative Pre-trained Transformers and BERT-based models, and integrated various classifiers. Throughout our work, we have focused on strategically integrating these components to ensure that each one enhances the accessibility of our sentiment classification pipeline. This concerted effort resulted in our sentiment classification approach that operates without the need for manual labeling, expert knowledge in tuning and data annotation, or extensive computing resources. We tested our approach on datasets from the retail, service, and cultural industries, where it consistently demonstrated strong performance. Our experiments also confirmed that the deployment of our approach does not require substantial computing resources and eliminates the need for manual labeling. This makes our approach highly accessible and allows more users to benefit from advanced sentiment classification techniques.

Furthermore, we believe this work not only provides a practical solution to the challenges of sentiment classification but also paves the way for applying our methodology in other fields. This expansion could enable a broader audience to benefit from machine learning technologies, which is a primary direction for our future work. By simplifying the application of sophisticated machine learning models, we aim to foster wider adoption and understanding, ultimately enhancing decision-making processes and insights across various sectors.


\bibliographystyle{IEEEtran}
\bibliography{references}

\end{document}